\documentclass[letterpaper, 10 pt, conference]{ieeeconf}  

\IEEEoverridecommandlockouts                              

\usepackage{amsmath} 
\usepackage{amssymb}  
\usepackage{siunitx}
\usepackage{pifont}
\newcommand{\xmark}{\ding{55}}%

\usepackage{booktabs}
\usepackage{tikz}

\usepackage{float}
\usepackage{lipsum}
\title{\LARGE \bf
Locomotion Policy Guided Traversability Learning using Volumetric Representations of Complex Environments
}

\author{Jonas Frey, David Hoeller,  Shehryar Khattak, Marco Hutter
\thanks{J. Frey, D. Hoeller, S. Khattak and M. Hutter are with the Department of Mechanical and Process Engineering, ETH Zürich, 8092 Zürich, Switzerland \{jonfrey, skhattak, dhoeller, mahutter\}@ethz.ch. D. Hoeller is also associated with NVIDIA.
This work was supported by the Swiss National Science Foundation (SNSF) through project 166232, 188596, the National Centre of Competence in Research Robotics (NCCR Robotics), and the European Union's Horizon 2020 research and innovation program under grant agreement No.780883.
}
}
\newcommand{\etal}{et al.}
\newcommand{\ra}[1]{\renewcommand{\arraystretch}{#1}}

\begin{document}

\maketitle
\thispagestyle{empty}
\pagestyle{empty}

\begin{abstract}
Despite the progress in legged robotic locomotion, autonomous navigation in unknown environments remains an open problem. Ideally, the navigation system utilizes the full potential of the robots' locomotion capabilities while operating within safety limits under uncertainty. The robot must sense and analyze the traversability of the surrounding terrain, which depends on the hardware, locomotion control, and terrain properties. It may contain information about the risk, energy, or time consumption needed to traverse the terrain. To avoid hand-crafted traversability cost functions we propose to collect traversability information about the robot and locomotion policy by simulating the traversal over randomly generated terrains using a physics simulator. Thousand of robots are simulated in parallel controlled by the same locomotion policy used in reality to acquire 57 years of real-world locomotion experience equivalent. For deployment on the real robot, a sparse convolutional network is trained to predict the simulated traversability cost, which is tailored to the deployed locomotion policy, from an entirely geometric representation of the environment in the form of a 3D voxel-occupancy map. This representation avoids the need for commonly used elevation maps, which are error-prone in the presence of overhanging obstacles and multi-floor or low-ceiling scenarios. The effectiveness of the proposed traversability prediction network is demonstrated for path planning for the legged robot ANYmal in various indoor and natural environments. --- Video: youtu.be/GGQ72tbAq0E

\end{abstract}

\section{INTRODUCTION}
\begin{figure}[ht]
\includegraphics[width=\linewidth]{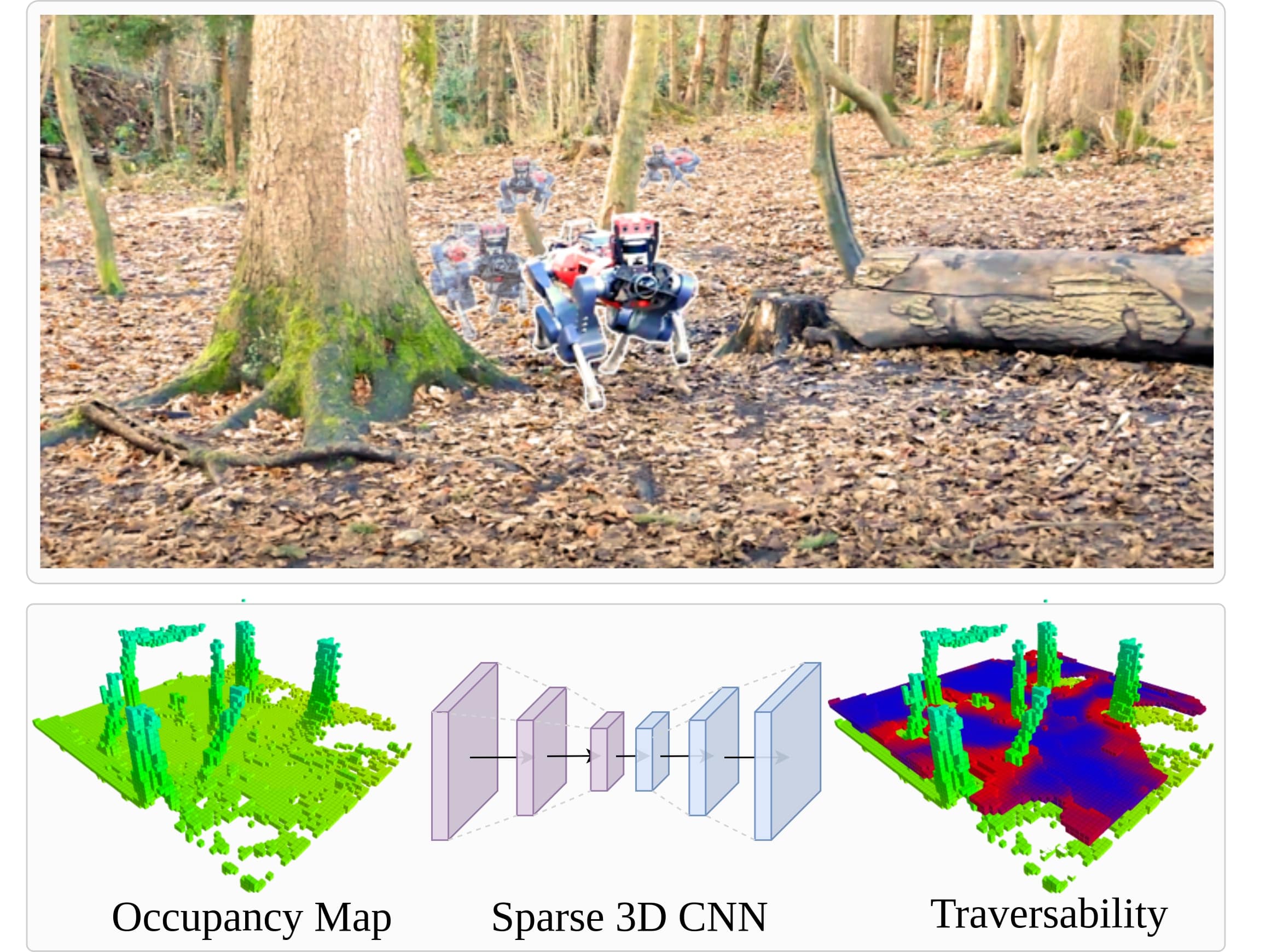}
\caption{The legged robot ANYmal is deployed in a forest. A robot-centric occupancy map is generated using two onboard LiDAR sensors. Our real-time capable proposed sparse 3D neural network predicts the traversability of the surrounding terrain in 3D, tailored to the deployed locomotion policy. Blue indicates risk-free traversable, while red indicates traversable under high-risk.}
\vspace{-0.5cm}
\end{figure}
Humans and animals have the remarkable ability to estimate where they can go safely and efficiently.
The capability to quickly assess one's surrounding is an essential selection criterion for survival in a variety of scenarios e.g., when a mouse has to escape from a cat into a narrow hole in the wall.
Similar constraint-aware terrain assessment capabilities are desirable for autonomous mobile robot deployment in complex unstructured environments. 
This assessment of the surrounding terrain can be used to navigate safely from a start to a goal position, during exploration, or to build a better semantic understanding of the scene. 
A specific terrain is associated with a cost, which depends on the capability of the robotic system, locomotion policy, and the structure of the deployment environment. 
The traversability assessment is performed based on multiple application-dependent objectives, which may include: Desired goal position, time and risk minimization, collision avoidance, safety margins, or energy efficiency~\cite{Panagiotis2013}. 

To acquire information about its surrounding, a robot is equipped with a suite of exteroceptive sensors.
Typically, geometrical information about the environment is accumulated in a map representation by fusing depth measurements from LiDARs or depth cameras.
For this, ground robots typically utilize a 2.5D elevation map representation due to its computational efficiency, in which each cell within a regularly sampled grid at a fixed resolution stores the estimated height value~\cite{Fankhauser2014RobotCentricElevationMapping}. 
During path planning for ground robots the elevation map can be used to restrict the search space, given that the base height of the robot is constrained by the underlying terrain geometry. 
This reduces the planning time given that not all poses in 3D space have to be considered for path planning. 

However, this 2.5D elevation map representation cannot represent multi-floor environments adequately~\cite{Philipp17}, as without additional semantic understanding, integration of depth measurements coming from over-hang obstacles or the ceiling corrupts the elevation map and leads to sub-optimal path planning.
Transitioning to a more expressive 3D voxel occupancy representation~\cite{Helen17} of the environment resolves the aforementioned problems, but requires new methods for efficient traversability assessment.
To this end, this work proposes a robot-centric traversability estimation module, which predicts the traversability of the terrain, while being aware of the robots' specific locomotion capabilities.
As an input, the volumetric occupancy representation is provided, which captures geometric information sufficient to avoid positive obstacles, negative obstacles, and steep slopes. 
To exploit the sparsity of the underlying problem, we use a sparse 3D convolutional neural network (CNN) to estimate the terrain traversability in real-time. 
The network is trained on data gathered within a highly-efficient physics simulator, where the success rate of executing different local motion commands on randomly generated terrain is evaluated inspired by the work of~\cite{Omar17, Guzzi20, Bowen21}. 
We deploy the same locomotion policy in simulation as on the real robots to achieve successful sim-to-real transfer. 
The main contributions of the presented work are:
\begin{itemize}
    \item Predicting terrain traversability using volumetric environment representations to allow for motion planning in complex environments with over-hanging obstacles.
    \item Application of a sparse 3D CNN trained exclusively on simulated data for traversability estimation.
    \item Creating a framework for large-scale traversability experience collection within a physics simulator.
    \item Deployment in real complex and natural environments and showcasing path planning application for ANYmal.
\end{itemize}

\section{RELATED WORK}
\begin{figure*}[t]
\centering
\includegraphics[width=1\textwidth]{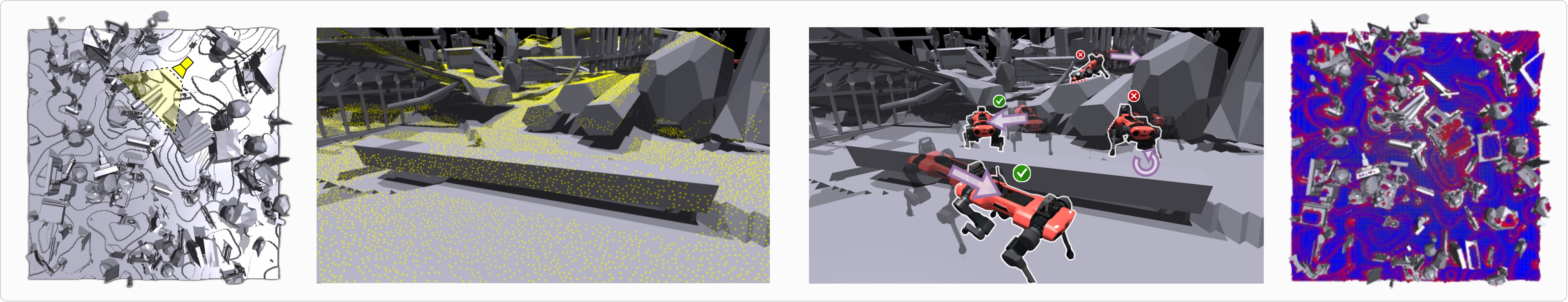}
  \caption{Collecting traversability information in simulation (left to right): Birds-eye-view of randomly generated terrain mesh; Zoomed-in view of collision-free starting points in yellow; Multiple simulated ANYmal robots executing motion commands; Simulated traversability estimate color-coded.}
 \label{fig:terrain_gen}
 \vspace{-0.5cm}
\end{figure*}
In this section, we provide an overview of traversability estimation methods deployed for ground robots.
We categorize the approaches based on their input representation into visual and geometrical methods.
Additionally, we provide an overview of the most relevant work for deep learning on volumetric voxel data.

\subsection{Visual Traversability Estimation}
Images contain rich geometrical and semantic information about the environment. 
Teams within the Darpa LAGR project~\cite{jackel06} deployed self-supervised learning approaches to classify from images the affordance necessary to traverse with wheeled robots outdoor. 
To overcome the issue of limited labeled data Otsu~\etal~\cite{Otsu16} train separate classifiers on vibration and visual data to classify terrain. Both classifiers are adapted using co- and self-training to improve the performance. 
Wellhausen~\etal~\cite{Wellhausen18} train a CNN to estimate terrain properties with a supervision signal acquired by force-torque sensors in a legged robot's feet. The supervision signal is extracted in hindsight in a self-supervised manner and associated with previously captured images of the terrain.
In conclusion visual data contains rich information to estimate terrain friction, terrain roughness, foothold quality, vegetation, support surfaces, and obstacles.
Extraction of this information under sensor noise, motion blur, bad lighting conditions remains a challenging problem and collecting labeled high-quality data for supervised learning methods is expensive. 
Further, using synthetic camera data is often prohibited by the sim-to-real gap induced by the mismatch between the simulated camera data and data encountered by a robot during its mission~\cite{jonas22}.

\subsection{Geometrical Traversability Estimation}
Apart from cameras, most mobile robots are equipped with LiDAR and depth cameras to acquire geometrical information. 
This data is less rich in semantic information but easier to process and simulate.
Individual depth measurements can be used for precise localization and fused into different representations including pointclouds, 2.5D elevation maps~\cite{Fankhauser2014RobotCentricElevationMapping, TakaElevationMapping} and 3D voxel-based occupancy maps. The most common frameworks to create the latter are Voxblox~\cite{Helen17}, Octomap~\cite{Hornung13}, Fiesta~\cite{Luxin19} or implementations build around the VDB representation~\cite{Museth13}.

Wermerlinger~\etal~\cite{Wermelinger16} use an elevation map and identify the traversability for a legged robot based on hand-crafted heuristics such as slopes, terrain roughness, and steps. 
Different risk sources are assessed from elevation data~\cite{Amanda20,FAN21}. The individual risks are stored in a Multi-Layer Traversability Map (MLT), which by superposition of all layers, is used to estimate the total risk for planning. 
Wellhausen~\etal~\cite{Lorenz21} train a CNN to estimate good footholds for a legged robot from an elevation map. The training dataset of good footholds is hand-labeled.

In contrast to previous work, Kruesi~\etal~\cite{Philipp17} use pointclouds to estimate the static terrain traversability for a wheeled robot to achieve safe motion planning.
Methods such as GBPlanner2~\cite{kulkarni2021} illustrate the benefit of a volumetric environment representation for exploration and global path planning. 
The planner is primarily developed for aerial robots, but when adapted to legged systems, the ground surface estimation is solved based on heuristics and traversability not considered for planning. 
All of the previous geometrical methods rely on hand-crafted heuristics and human expertise to evaluate traversability.

In contrast, Chavez-Garcia~\etal~\cite{Omar17} simulate a wheeled robot executing local motion commands and gather traversability information in simulation. 
They train a CNN to classify the traversability of elevation maps.
Guzzi~\etal~\cite{Guzzi20} extend this concept for legged robots and, additionally to the robot-centric elevation map provide a relative target pose to estimate the time and probability of reaching the target. 
Lastly, Yang~\etal~\cite{Bowen21} follow a similar procedure to simulate the traversability of a given terrain for a legged robot. 
Information about the time, risk, and energy needed for a robot to execute a local motion command is collected in simulation. 
A CNN network is trained using this data and predicts the collected quantities similar to~\cite{Guzzi20}. 
Our approach follows the idea of~\cite{Guzzi20} to simulate the robot to acquire locomotion policy-guided traversability information. 
All previous work rely on elevation maps, which are error-prone in multi-floor or overhanging obstacles scenario, to estimate traversability during the mission. 
We instead use a more expressive occupancy input representation resolving this failure mode. 

\subsection{Voxel-based Deep Learning}
Predicting traversability from 3D voxel-based occupancy data requires computational efficient methods mitigating the underlying cubic complexity by exploiting sparsity. 
Wang~\etal~\cite{Peng17} develop a neural network to operate on an octree volumetric data structure for shape analysis, restraining computations on occupied voxel octants.
Sparse 3D CNNs restrain computation to occupied voxels~\cite{Graham17}. 
These networks have been generalized to higher dimensions by Choy~\etal~\cite{choy19}. 
We adopt the sparse encoder-decoder network of Gwak~\etal~\cite{Gwak20} for semantic segmentation to predict traversability. 
The network utilizes sparse generative convolutions and pruning with sequential up-sampling and pruning voxels in 3D.
\section{APPROACH}
\label{sec:app}
\subsection{Problem Definition:}
\label{subsec:app_pro}

Given an environment representation and the internal state of the robot we assess if the robot is capable to reside over a given terrain.
The environment is represented by an occupancy voxel map $\mathbf{O}(x,y,z) \in \{0,1\}$ with a resolution of $d_{voxel}$, where each voxel is either labeled as free(0) or occupied(1).
The state of the robot~$\mathbf{s}$ is composed of its position $\mathbf{p}~=~(x,y,z)^T$, heading~$\alpha$, and locomotion policy~$\pi_\theta$,
\begin{equation}
    \mathbf{s} = (\mathbf{p},\alpha,\pi_\theta)^T
\end{equation}
To store the traversability information associated with a terrain we introduce the notion of a traversability cost tensor~$\mathbf{T}$. 
The tensor stores the traversability cost at a spatial location, for a set of robot configurations and motion commands~$a$. 
We simplify the robot configuration to solely include the heading parameter $\alpha$ and discard velocities, individual joint positions and base pitch and roll, given that the traversability assessment is less dependent on these fast varying variables under nominal operating conditions of the studied legged robot ANYmal.
Including the robot configuration and motion command is motivated by the fact that traversing up or down a slope imposes varying risk and e.g. within tight corridors only a subset of actions can be executed collision-free.
The traversability cost tensor allows to capture this information adequately and measures traversability in a range of 0 to 1, where 0 is untraversable and 1 risk-free traversable.
\begin{equation}
    \mathbf{T}_{\pi_\theta}(x,y,z,\alpha_j, a_j) \in [0,1]
\end{equation}
Our objective is to find the function $f_{trav}$ which predicts each element within the traversability cost tensor $\mathbf{T}$, given the occupancy map representation $\mathbf{O}\in\mathcal{O}$ and state of the robot $\mathbf{s}\in\mathcal{S}$. 
\begin{equation} \label{equ:trav_function1}
\begin{split}
    \mathbf{T}_{\pi_\theta}(x,y,z,\alpha_i, a_j) = f_{trav}( \mathbf{s}, \mathbf{O} ) \\
    f_{trav}: \quad \mathcal{S} \times \mathcal{O} \rightarrow \mathbb{R}_{[0,1]}
\end{split}
\end{equation}

\subsection{Simulating Traversability:}
\label{subsec:app_sim}
\begin{figure*}[t]
\centering
\includegraphics[width=\linewidth]{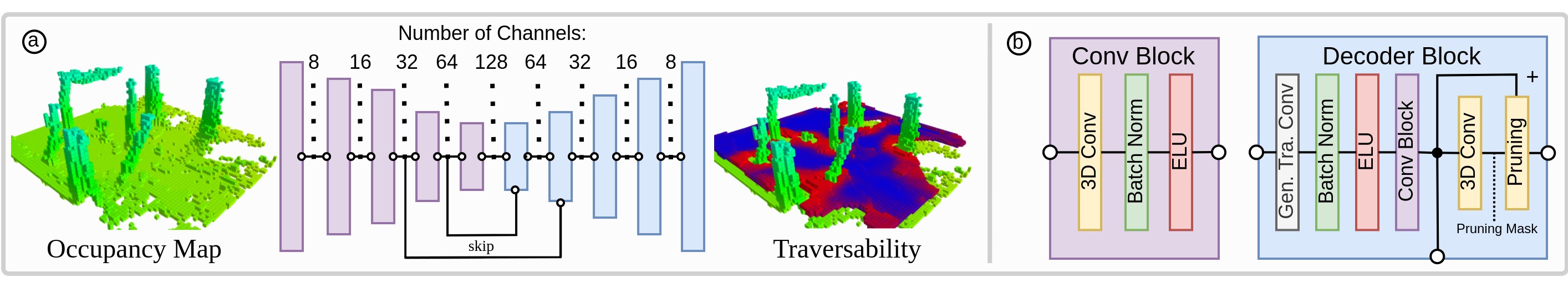}
\caption{(a) Simplified Sparse Encoder-Decoder 3D CNN Architecture, (b) Block Structure (for further detail refer to Section~\ref{subsubsec:net} and~\cite{Gwak20})}
\vspace{-0.5cm}
\label{fig:architecture}
\end{figure*}
The traversability cost is evaluated by simulating locomotion over randomly generated terrain in NVIDIA's IsaacGym~\cite{Makoviychuk21}. 
To achieve accurate sim-to-real transfer the robot is controlled by the same locomotion policy later deployed on the real robot. 
In this work, the traversability cost is defined as the success rate of reaching the desired goal pose, which is of main interest for autonomous navigation. 
The goal pose is determined by the motion command with respect to the robots starting pose. 
Evaluation of multiple trials per motion command under domain randomization results in a robust geometric traversability risk assessment.
We introduce the terrain generation procedure in~(Sec.~\ref{subsub:tg}), followed by the traversability data collection in~(Sec.~\ref{subsub:dc}). 
\subsubsection{Terrain Generation}
\label{subsub:tg}
We aim to generate a large and diverse terrain dataset covering the domain of terrains that the robot may encounter during deployment.
Acquisition of a large amount of experience has shown to increase generalization performance for various tasks, including locomotion~\cite{miki2021locomotion} and traversability estimation~\cite{Bowen21}. 
The ground structure of the terrain map is generated using Perlin noise with either a smooth surface or height-varying steps, following the commonly used terrain generation for training locomotion policies~\cite{miki2021locomotion, Rudin22}. 
To generate a complex environment with a variety of static obstacles, multiple objects with a diameter between \SI{0.1}{m} and \SI{6}{m} are randomly spawned in the map. 
These objects were originally generated by human artists to create 3D worlds for computer games. 
In total between 300 and 1000 objects are spawned with a random location and orientation. 
The spawning probability of each object is inversely scaled with the diameter of its axis-aligned bounding box leading to more smaller objects.
Each spawned object's size is randomly scaled with a factor of $0.5$ and $1.5$. 
With a probability of \SI{90}{\percent} the spawned object is aligned with the ground surface and otherwise spawned uniformly sampled between \SI{0}{m} and \SI{3}{m} above the ground resulting in overhanging obstacles. One example of a \SI{32}{m} $\times$ \SI{32}{m} terrain patch is shown in Figure~\ref{fig:terrain_gen}.

\subsubsection{Traversability Data Collection}
\label{subsub:dc}

An initial set of robot starting poses is evaluated by uniformly sampling points on the generated terrain mesh. 
We evaluate if the robot can be spawned collision-free with nominal joint configuration above the sampled point. 
This is conducted for all headings $\alpha_k$ sampled with $10^\circ$ resolution, resulting in a total of 36 possible start headings per point.
The robots' pitch and roll base orientation are aligned with the underlying mesh. 
The base height is adjusted such that all feet are in proximity to the mesh surface.
Only a single admissible robot starting pose for each heading is further evaluated per voxel with the same spatial resolution as the traversability cost tensor~$\mathbf{T}$. 
Finally, to verify that the start position is safe for the robot to be initialized, \SI{2}{s} of standing is simulated within the physics simulator with the current locomotion policy. 
If the robot is capable to remain at the same position successfully without collisions, the pose is labeled as a valid starting pose.
Determining good starting poses is essential to reduce failure during traversability data collection due to a bad initialization of the robot. 
In addition, data collection is only performed for a reduced subset of initial starting poses in order to reduce the compute requirements. 

For each of the valid spawn pose, $n_{total}=10$ trials of each motion command $a_i$ are executed.
In particular, 4 translation and 2 rotation motions are performed.
The translation distance is set to \SI{40}{cm} along four cardinal directions (forward, backward, left, right) and the rotation is given by a $\pm$\SI{45}{\degree} yaw.
The selected motion commands determine the relative pose offset with respect to the starting pose.
During execution the current pose offset of the robot with respect to the goal is used to compute the velocity command tracked by the locomotion policy. 
Four different trajectories with start (transparent) and goal poses (dense) with purple arrows indicating the motion command are shown in Figure~\ref{fig:terrain_gen}. 
Each attempt results in a robot trajectory, which is evaluated for success based on foot contacts, base orientation, environment collision, time to reach the goal and distance to the goal.
A maximum execution time of \SI{4}{s} is set for each attempt. 
In Figure~\ref{fig:terrain_gen} the forward and side-ward moving robots reach their goal successfully (green check mark), whereas the other robots fail to execute the desired motion command given a collision with the environment (red cross). 
The overall traversability cost is given by the number of successful attempts $n_{suc}$ divided by total attempts $n_{total}$.

We choose to use small pose offsets, such that a motion command can be correlated with the terrain surrounding the starting pose. 
Therefore the outcome of each trajectory is solely associated with the single starting entry of the traversability cost tensor, which is indexed by the starting robot position.
Evaluating longer trajectories leads to the problem of determining which terrain part is responsible for the failed execution.
Additionally, it is possible to increase the safety limits by adjusting the size of the used collision bodies.
For each attempt the robot base mass and friction coefficient are randomized. 
The procedure to evaluate a trajectory on success can be further adapted depending on the application, which may include time, number of steps, or the robots’ configuration.
A sample of the collected terrain traversability cost is color-coded in Figure~\ref{fig:terrain_gen}, where blue indicates risk-free traversable and red traversable under high risk. 

\subsection{Sparse 3D Traversability Prediction Network}
\label{subsec:app_spa}

\subsubsection{Input}
We train a sparse 3D CNN in a supervised manner to predict traversability information in real-time. 
Offline, for each terrain patch, the mesh is converted to a voxel occupancy map with a resolution of \SI{10}{cm} using ray tracing. 
Every voxel containing a triangle is labeled as occupied.
The occupancy map generated during the mission provides fused information of all past observations while functioning as an abstraction layer.
The previously computed robot starting poses are used to crop a robot-centric training example from the larger terrain patch. The robot's base position is at the center of the extracted volume. The volume is rotated to align with robot's base yaw, while the z-axis is gravity aligned. 
This allows to implicitly provide the network the robot's pose, while the inclination of the terrain remains unchanged. 
We choose to use a window size of \mbox{\SI{8}{m} $\times$ \SI{8}{m} $\times$ \SI{4}{m}} or respectively a volume size of \mbox{$80\times80\times40$} voxels at a \SI{10}{cm} resolution. 
The sparse 3D CNN is implemented using the Minkowski Engine~\cite{choy19}.

\subsubsection{Network}
\label{subsubsec:net}
We adapt the generative sparse detection network developed for semantic segmentation~\cite{Gwak20}. 
The network consists of an encoder-decoder structure with skip connections between the respective feature maps and is illustrated in Figure~\ref{fig:architecture}.
Each encoder block applies a sparse 3D convolution (stride=2, kernel=4), batch normalization, and exponential linear unit (ELU) as an activation function, which downsamples the spatial resolution by a factor of 2. 
The encoder generates feature maps with \mbox{[8, 16, 32, 64, 128]} channels. 
In contrast to the original network architecture skip connections are only applied in the two lower layers of the architecture. 
The intuition behind the choice of reducing the number of skip connections is discussed in Section~\ref{subsec:exp_training}.
The decoder spatially up-samples the feature maps to the original resolution using generative transposed convolutional layers. 
At each decoder level a pruning mask is predicted by a sparse convolutional layer (stride=1, kernel=1) to increase the sparsity and reduce the computation needed.
Only non-zero elements are propagated to the following layers. 
Furthermore, we tested applying our network in three different output configurations, providing different levels of information that can be used for path planning purposes later. 
In the simplest case only the \textit{total risk} is assessed, where the commands and robot configuration are marginalized within the traversability cost tensor resulting in a $c=1$ dimensional target.
The \textit{orientation risk} when marginalizing over the motion commands results in a $c=18$ dimensional traversability score with a \SI{10}{\degree} resolution when accounting for the symmetry of the robot.
Lastly, the \textit{c-bin motion direction risk} for which the relative motion direction of the applied action-heading combination, is used to bin them in $c$-equally sized motion direction bins. 

\subsubsection{Training}
The network is trained to minimize the reconstruction and traversability estimation loss.
The reconstruction loss is evaluated per decoder layer and is given by the weighted binary cross-entropy between predicted pruning mask and the respectively down-sampled target traversability cost tensor. 
The traversability loss is given by the mean squared error between the predicted and target traversability score. Here, the traversability loss is only assessed for correctly instantiated (true positive) points. 
ADAM with weight decay is used to optimize the network for 100.000 steps with a batch size of 8.
The learning rate is scheduled according to the \textit{1cycle} learning rate with a target learning rate of~$0.001$~\cite{Leslie17}.
During training at each network layer the correct-pruning mask is enforced for training stability. 

The occupancy map created based on the full available mesh data does not resemble the data sensed by a robot during its mission. 
Surfaces are mostly observed from limited viewpoints, under sensor noise and odometry errors, which degrade the quality of the occupancy map generated online during a mission.
We increase the robustness of the network using data augmentation. 
Firstly, only voxels are marked as traversable if a continuous traversable path exists, connecting the voxel to the starting robot position at the center. 
This is done by applying a flood-fill algorithm. 
The designed data augmentation exploits two underlying assumptions. 
Given the robots' size the traversability score is independent of the occupancy state of a voxel further away than \SI{1}{m} from it. Additionally, the traversability cost prediction should be invariant to minor surface imperfections, which commonly arise due to discretization effects when generating occupancy maps. 
These two assumptions allow us to randomly drop occupied voxels and add surface imperfections by randomly spawning voxels within the proximity of occupied voxels. 



\section{EXPERIMENTS}
\label{sec:exp}
\subsection{Training Results}
\label{subsec:exp_training}

\begin{figure}[ht]
\label{fig:hoeng}
\centering

\includegraphics[trim=0 100 0 200,clip,width=0.32\linewidth]{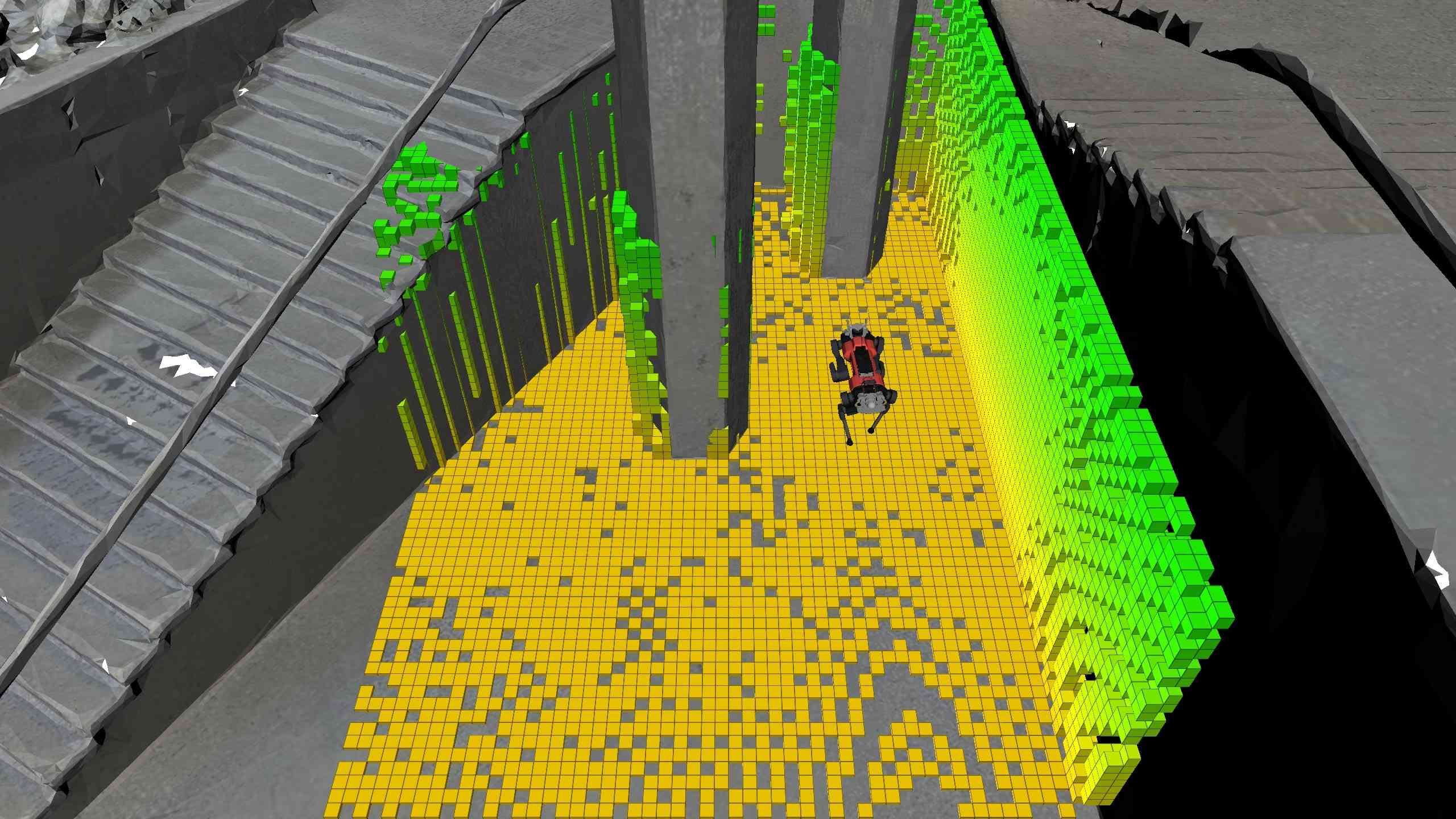}
\includegraphics[trim=0 100 0 200,clip,width=0.32\linewidth]{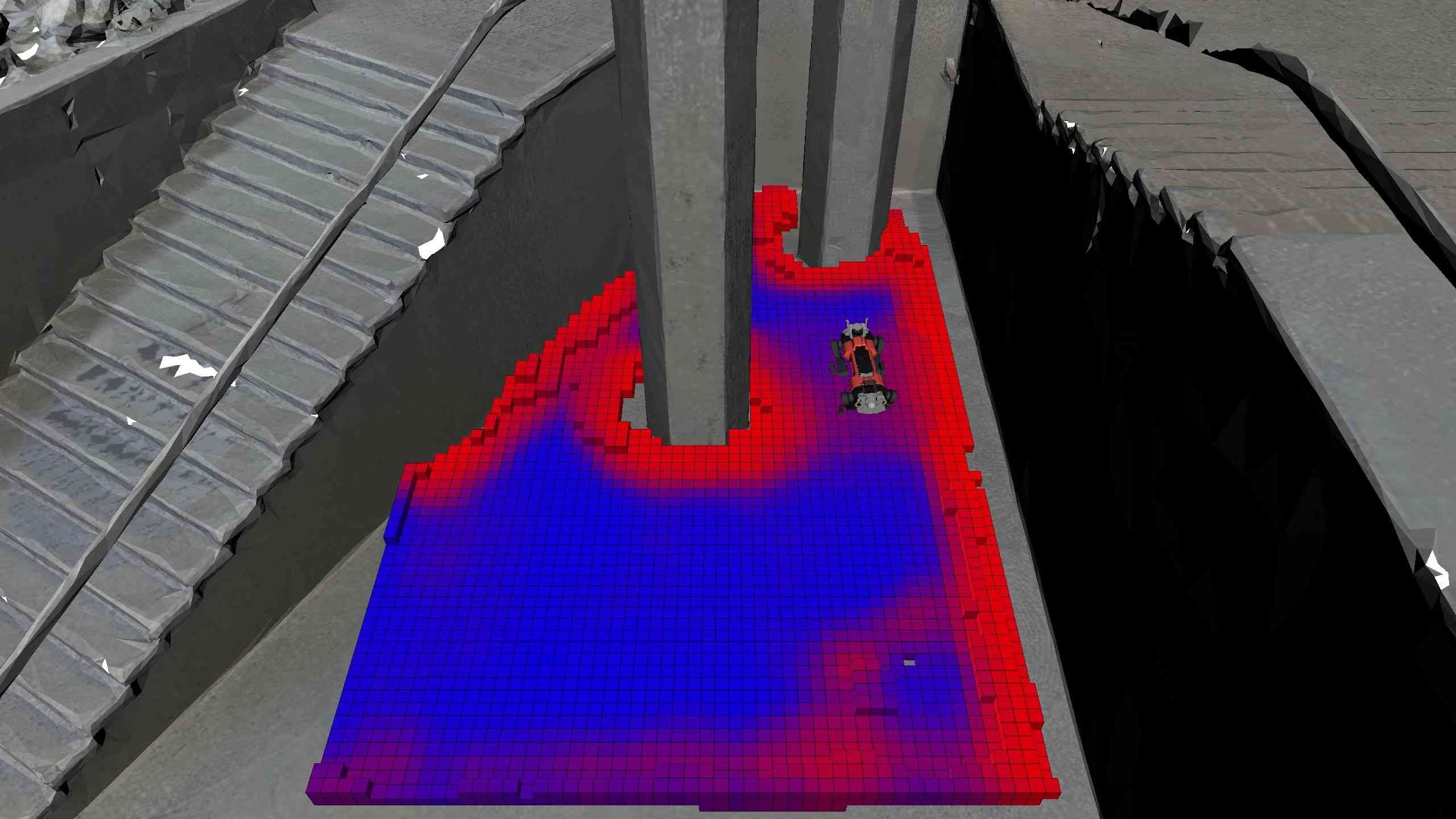}
\includegraphics[trim=0 100 0 200,clip,width=0.32\linewidth]{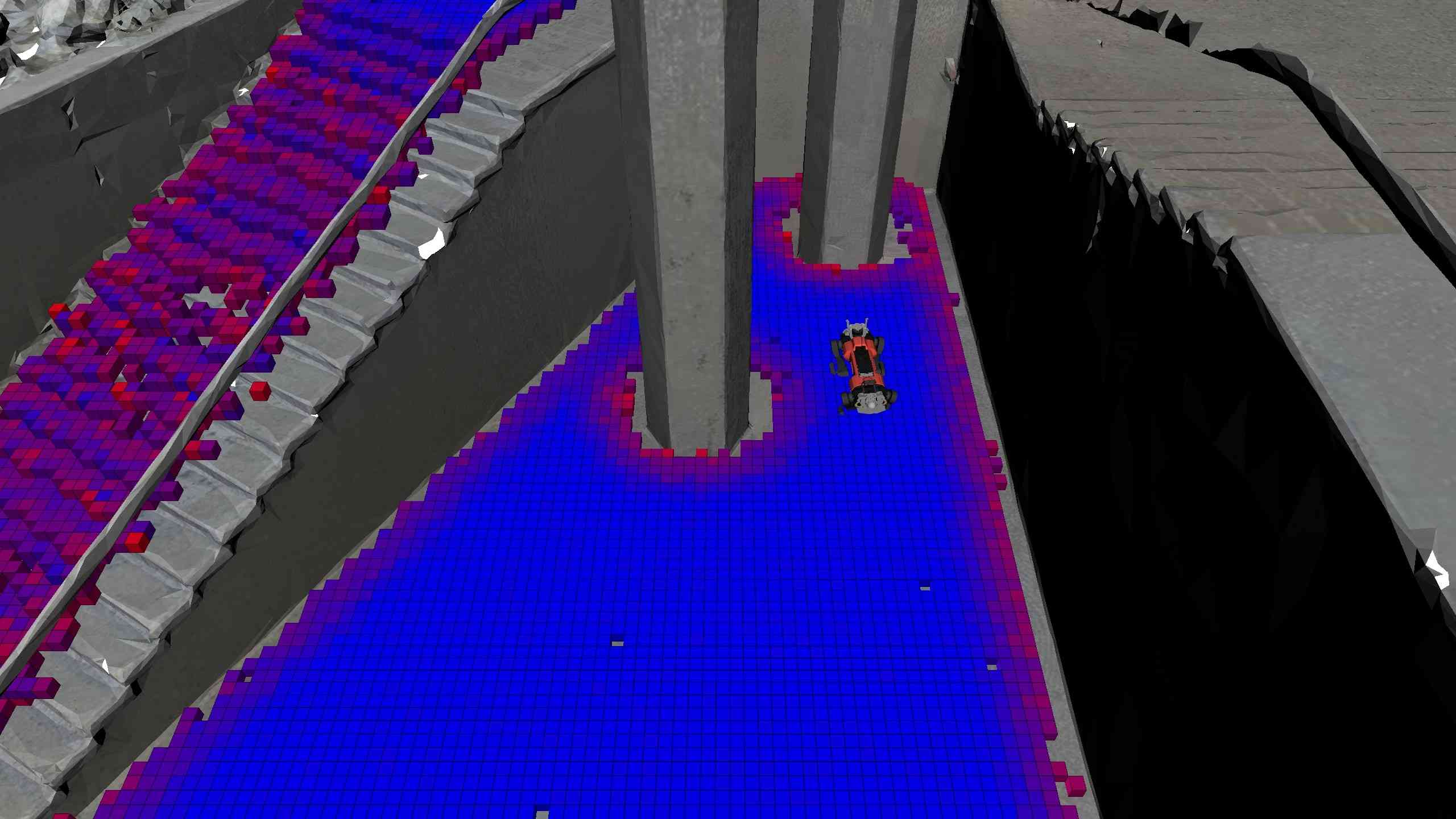}\\
\vspace{0.15cm}

\includegraphics[trim=0 100 0 200,clip,width=0.32\linewidth]{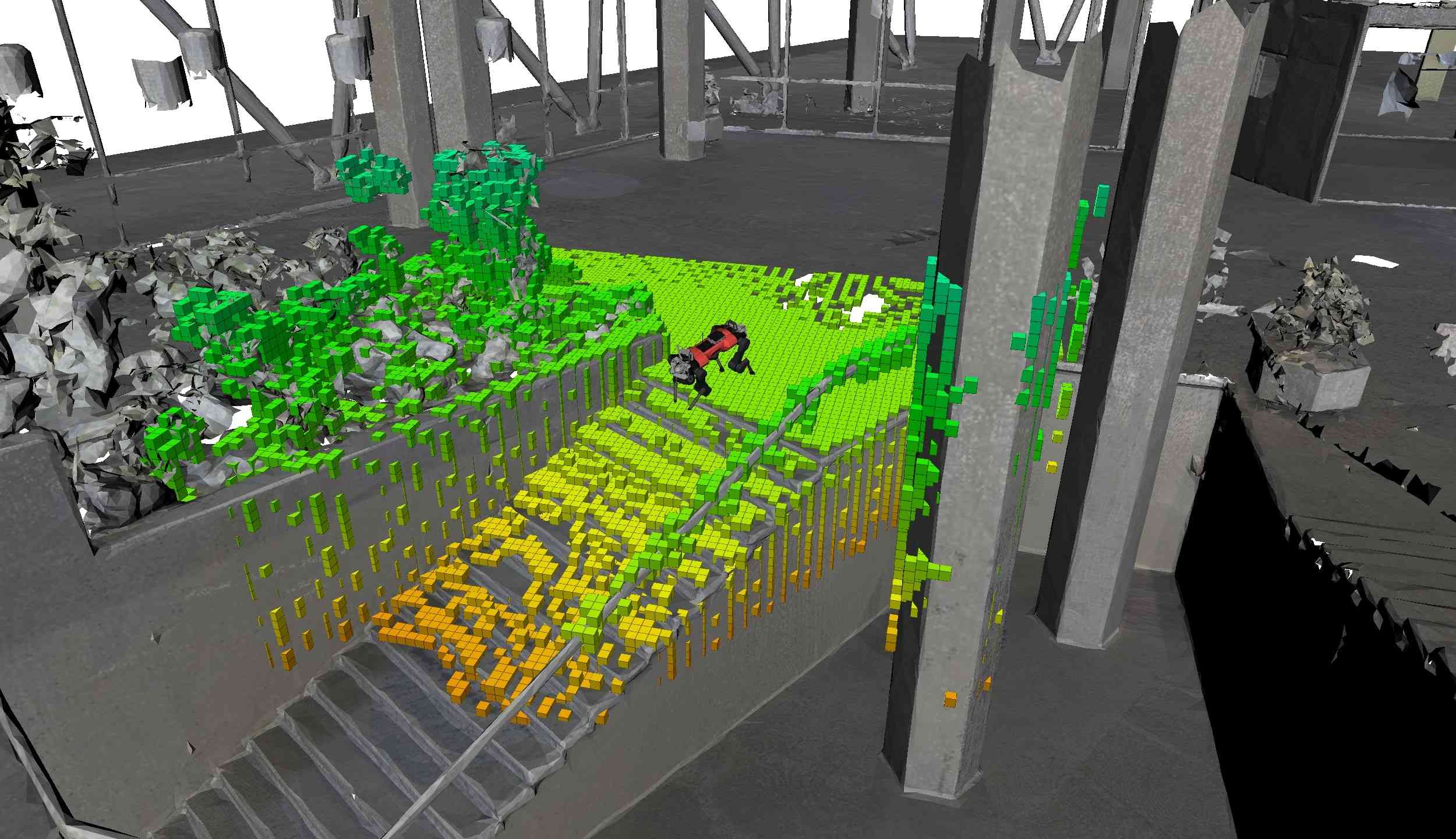}
\includegraphics[trim=0 100 0 200,clip,width=0.32\linewidth]{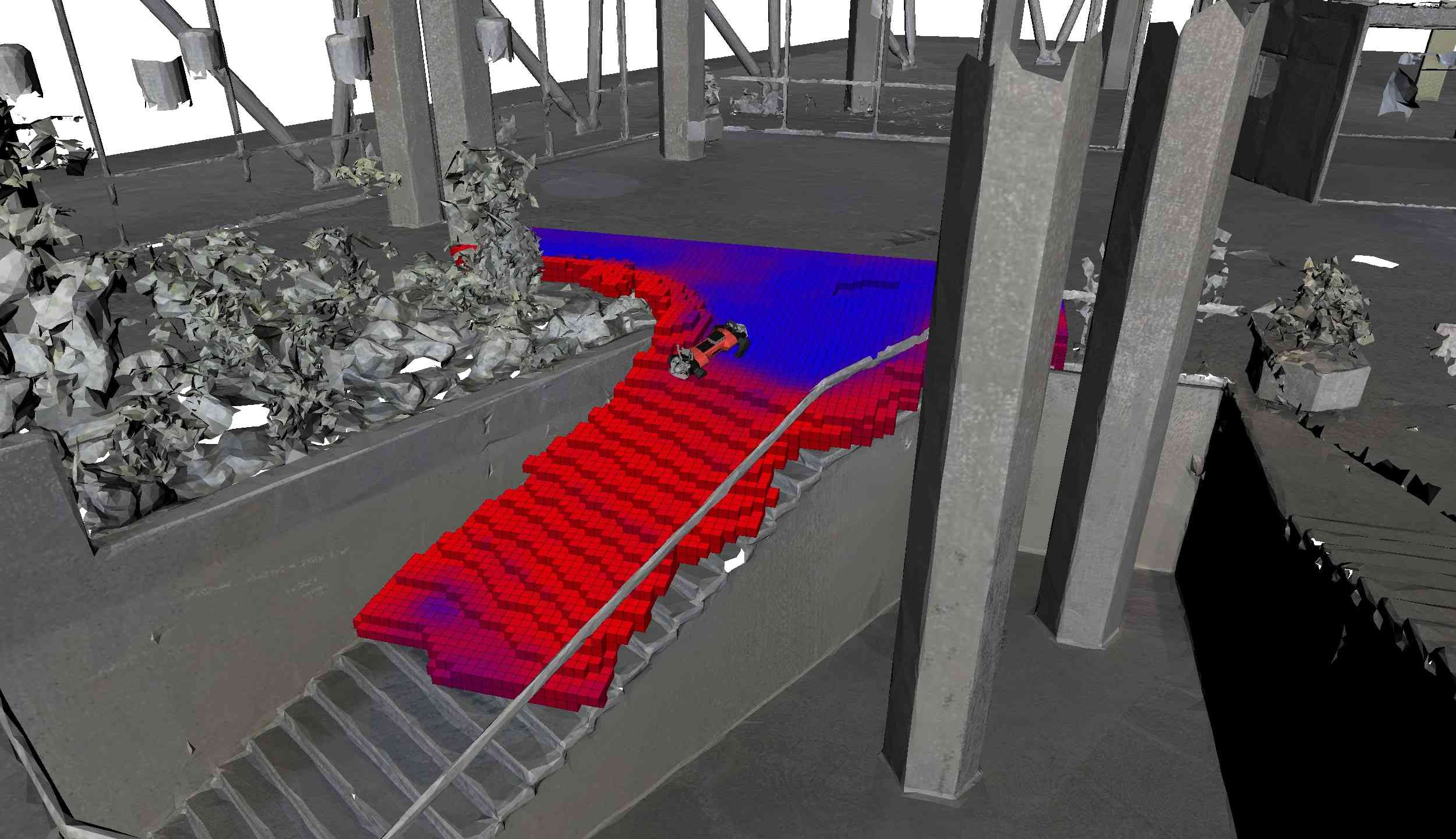}
\includegraphics[trim=0 100 0 200,clip,width=0.32\linewidth]{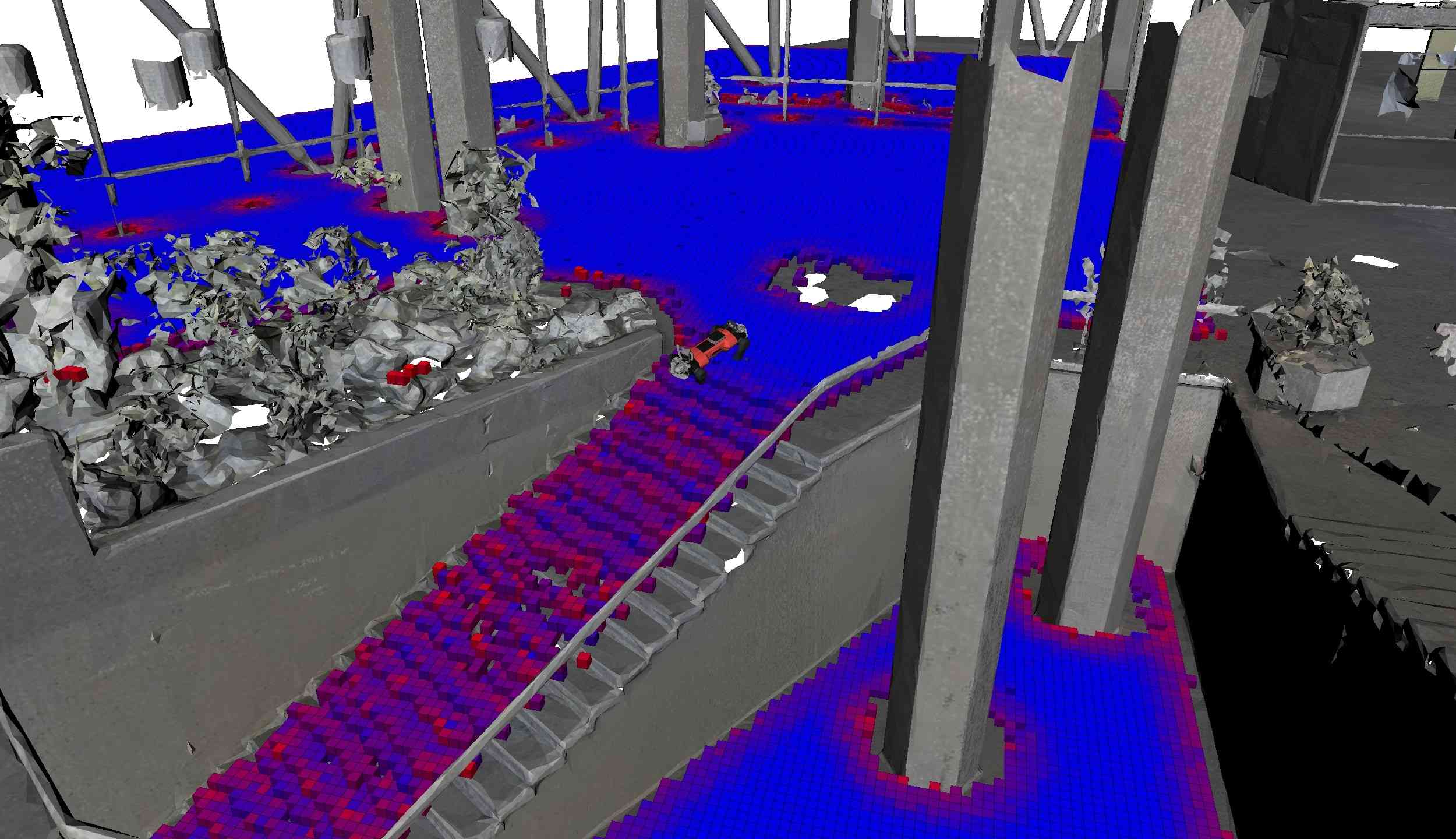} \\
\vspace{0.15cm}

\includegraphics[trim=0 100 0 200,clip,width=0.32\linewidth]{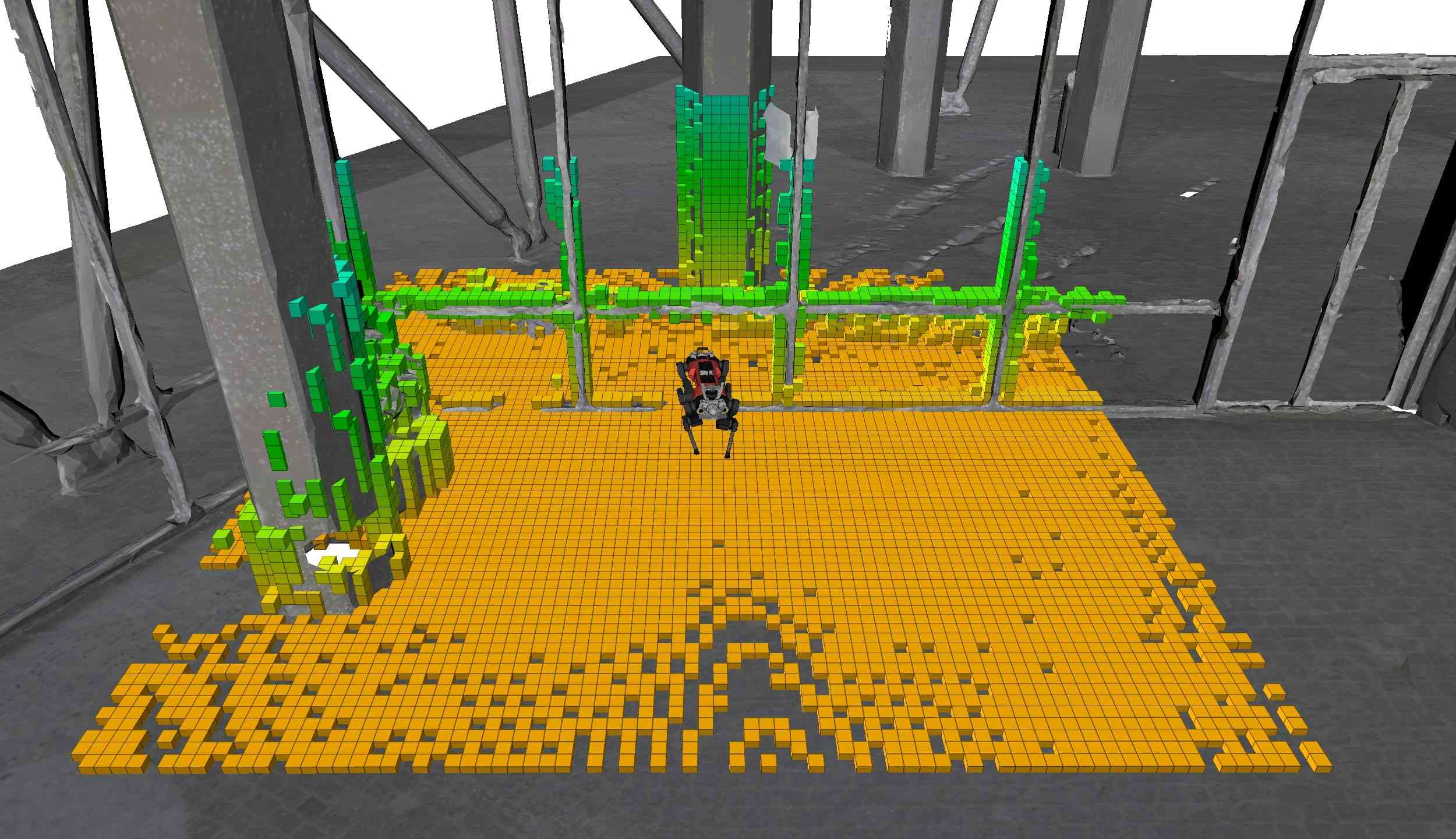}
\includegraphics[trim=0 100 0 200,clip,width=0.32\linewidth]{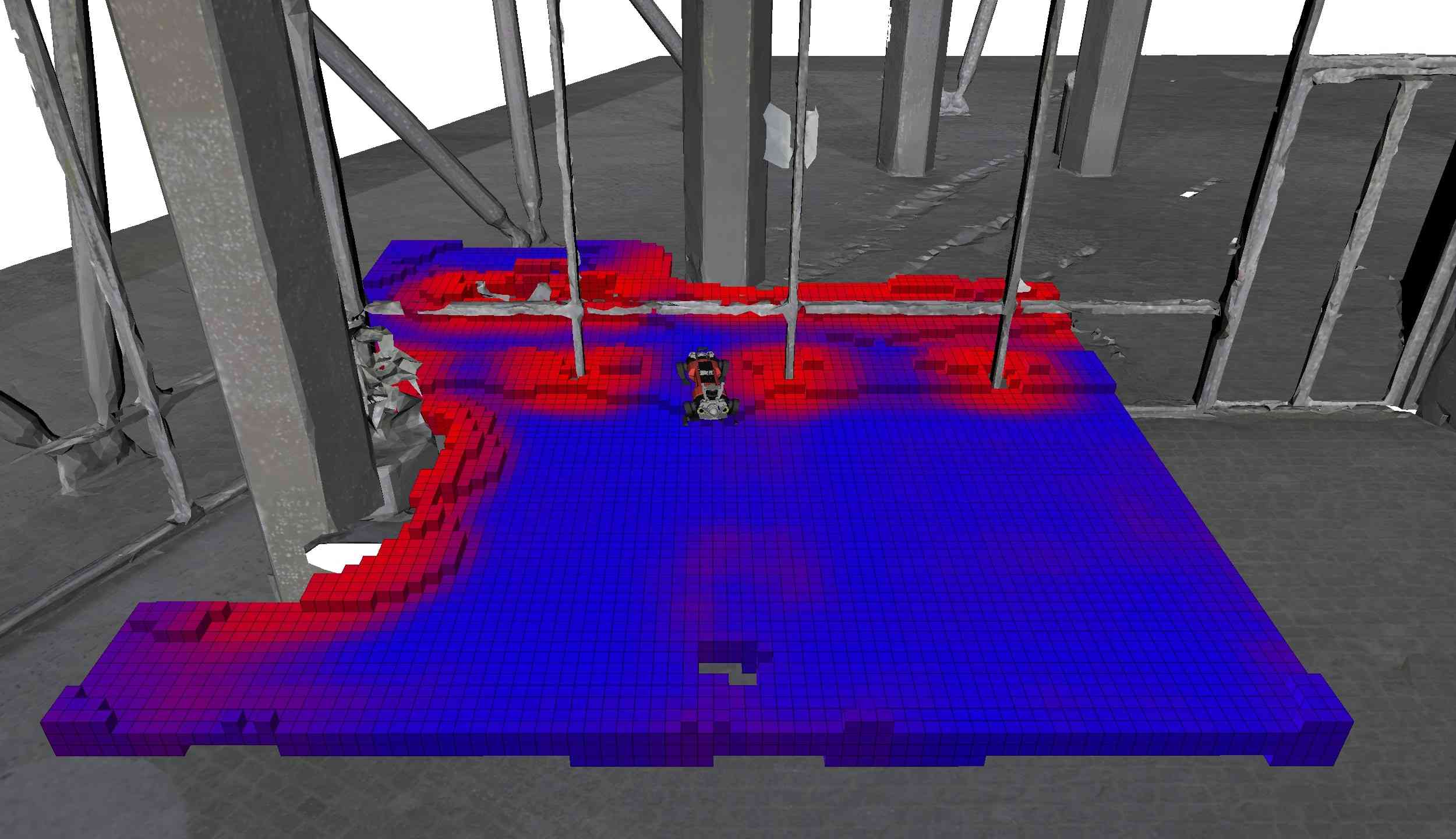} 
\includegraphics[trim=0 100 0 200,clip,width=0.32\linewidth]{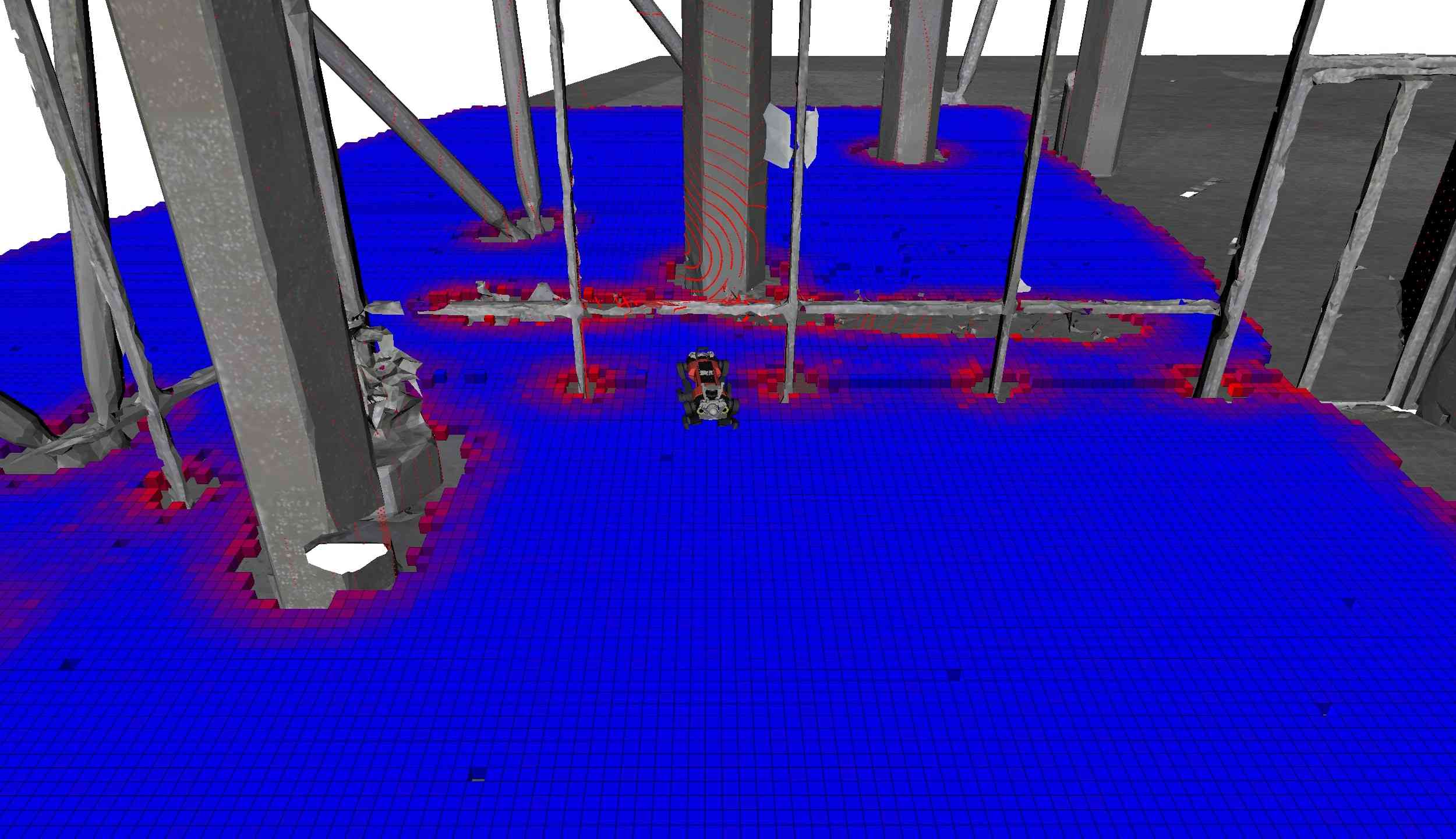}
\caption{ETH HPH building test dataset (left to right): Local Voxblox occupancy map, height is color-coded; Predicted \textit{total risk} traversability cost for local occupancy map (blue risk-free traversable, red traversable under high risk); Simulated \textit{total risk} traversability cost for full map-patch, (blue risk-free traversable, red traversable under high risk);}
\label{fig:hph}
\end{figure}

To train our proposed network, we created 20 training and 10 validation terrain patches with a size of \mbox{\SI{32}{m} x \SI{32}{m}} following the procedure explained in Section~\ref{subsec:app_pro}.
Separate object asset datasets are used for generating the training and validation terrain. 
These objects were generated by professional artists and are freely available on the Unity Asset Store. 
We simulate~4096 robots to collect traversability information in parallel facilitating an accumulated real-time speed up factor of~400. Per patch~2500 minutes wall time are spent collecting traversability using an  RTX2080TI and 18~CPU cores of a AMD EPYC 7763. In total 57~years of real-time equivalent traversability information was gathered. 

The test dataset is generated by applying the same traversability simulation procedure for two real-world meshes consisting of an underground mine in Switzerland (1 patch) and a multi-floor university building at the ETH Zurich (4 patches) shown in Figure~\ref{fig:hph}. 
The mine mesh was collected during a preparatory run of team CERBERUS~\cite{Tranzatto2022} for the DARPA Subterranean Challenge. 
All meshes were captured using a Leica RTC360 laser scanner.
While the training and validation datasets are drawn from the same distribution, the test dataset allows assessing performance on real-world data. 
In the absence of ground truth, the generated test dataset allows for assessing the network's capability to reproduce the simulated traversability cost from the limited geometrical information provided by the occupancy representation. 
In Section~\ref{subsec:exp_real} we experimentally validate how this traversability estimate transfers to the real-world and can be used for path planning. 
In Table~\ref{tab:train_val_test} we provide an overview of the generated datasets. 

\begin{table}[t]
\vspace{-0.2cm}
\centering
\ra{1.2}
\footnotesize
\setlength{\tabcolsep}{3pt}
\caption{Datasets Overview: Assets reefers to the number of assets gathered. Instances is the range of spawned assets per patch. }
    \begin{tabular}{lccccc} \toprule
     & Patches & Size [m] & Assets & Instances & Real \\ \midrule 
    \textbf{Train} & 20 & 32x32x16 & 300 & 300-1000 & \xmark \\ 
    \textbf{Val} & 10 & 32x32x16 & 200 & 300-1000 & \xmark \\
    \textbf{Test} & 5 & 32x32x16 & - & - &  \checkmark  \\\bottomrule
    \end{tabular}
\label{tab:train_val_test}
\vspace{-0.6cm}
\end{table}

The root-mean-squared error (RMSE) between the predicted traversability $\hat{y}$ and simulated traversability score $y$ is used to evaluate the performance. Not instantiated voxels are assumed to have a traversability of zero, and hence true negative (TN) voxels are not included when calculating the RMSE metric.
\begin{equation} \label{equ:trav_function2}
\begin{split}
    L_{RMSE} &= \sqrt{\frac{1}{N} \sum_{TP, FN, FP} (y-\hat{y})^2 } \\
    N &= |TP|+|FN|+|FP|
\end{split}
\end{equation}

\begin{table}[ht]
	\centering
	\ra{1.2}
	\footnotesize
	\setlength{\tabcolsep}{3pt}
	\caption{RMSE reported for different network architectures \textbf{M}, output channels \textbf{c}, and datasets (Train, Validation, Test)}
\begin{tabular}{lccclccclccclccclccc} \toprule
    & \multicolumn{3}{c}{$c=1$} && \multicolumn{3}{c}{$c=4$}  && \multicolumn{3}{c}{$c=18$} \\ 
    \cmidrule{2-4} \cmidrule{6-8} \cmidrule{10-12} \cmidrule{11-12} 
	& Tr & Val & Test && Tr & Val & Test && Tr & Val & Test \\ \midrule 
\textbf{M1} & 0.124 & 0.131 & 0.101 && 0.232 & 0.236 & 0.162 && 0.450 & 0.465 & 0.385 \\ 
\textbf{M2} & 0.119 & 0.126 & 0.097 && 0.221 & 0.226 & 0.155 && 0.438 & 0.450 & 0.379 \\ 
\bottomrule
\end{tabular}
\label{tab:compare_models}
\end{table}

During the training of the network, multiple hyper-parameters, including the learning rate, model capacity (number of channels), loss function hyper-parameters, data augmentations, and model architecture, were validated. 
We performed experiments for altering the number of skip-connections, which allow information forwarding between the encoder and decoder layers. 
Full skip-connectivity results in a large number of false positively classified voxels and overall worse performance (M1, Table~\ref{tab:compare_models}). 
Intuitively this can be explained by the fact that the decoder has to correctly predict the pruning mask for all forwarded features by the encoder. 
Only using skip connections in the two lower layers resulted in the lowest RMSE (M2, Table~\ref{tab:compare_models}). 
Furthermore, with the increase of task complexity (the number of output channels), the RMSE increases for all network architectures. 
This increase can be attributed to two reasons. First, the supervision signal includes more noise for a more complex task given that fewer robot trajectories are accumulated within a traversability estimate. 
Second, the model complexity is kept constant and a single pruning mask is predicted for all channels within the decoder.

Additionally, it can be noted that the performance on the test data is better than training data. 
We suspect that the real world data overall is an easier objective to estimate than the randomly generated terrain. 
This observation also demonstrates that the network is not overfitting to the training dataset. 
In Figure~\ref{fig:hoeng} we deploy ANYmal, simulated in Gazebo, on the mesh of the university building at the ETH Zurich. 
We simulate the LiDAR sensors and use their measurements to build a robot-centric occupancy map (left). 
The network prediction (middle) and simulated traversability cost for the full map are color-coded. 
The trained network is conservative on the stairs. 
In addition, the ground truth traversability estimate on the stairs is noisy but consistent over the full staircase.

\subsection{Real-World Deployment}
\label{subsec:exp_real}
We deployed ANYmal within various environments to verify the successful sim-to-real transfer. The robot is equipped with a front and back-facing RoboSense RS-BPearl LiDAR sensors.
First, we illustrate the advantage of using a more expressive occupancy map representation and compare it to an elevation mapping-based approach.
Second, we deploy the robot within a confined space to showcase the effectiveness of action aware traversability risk assessment. 
Last, we apply the network to data collected within a forest and an underground mine to showcase the effectiveness of our method for planning in complex real-world scenarios. 

\subsubsection{Overhanging Obstacles}
\label{subsubsec:exp_overhang}
\begin{figure}[ht]
\centering
\includegraphics[width=\linewidth]{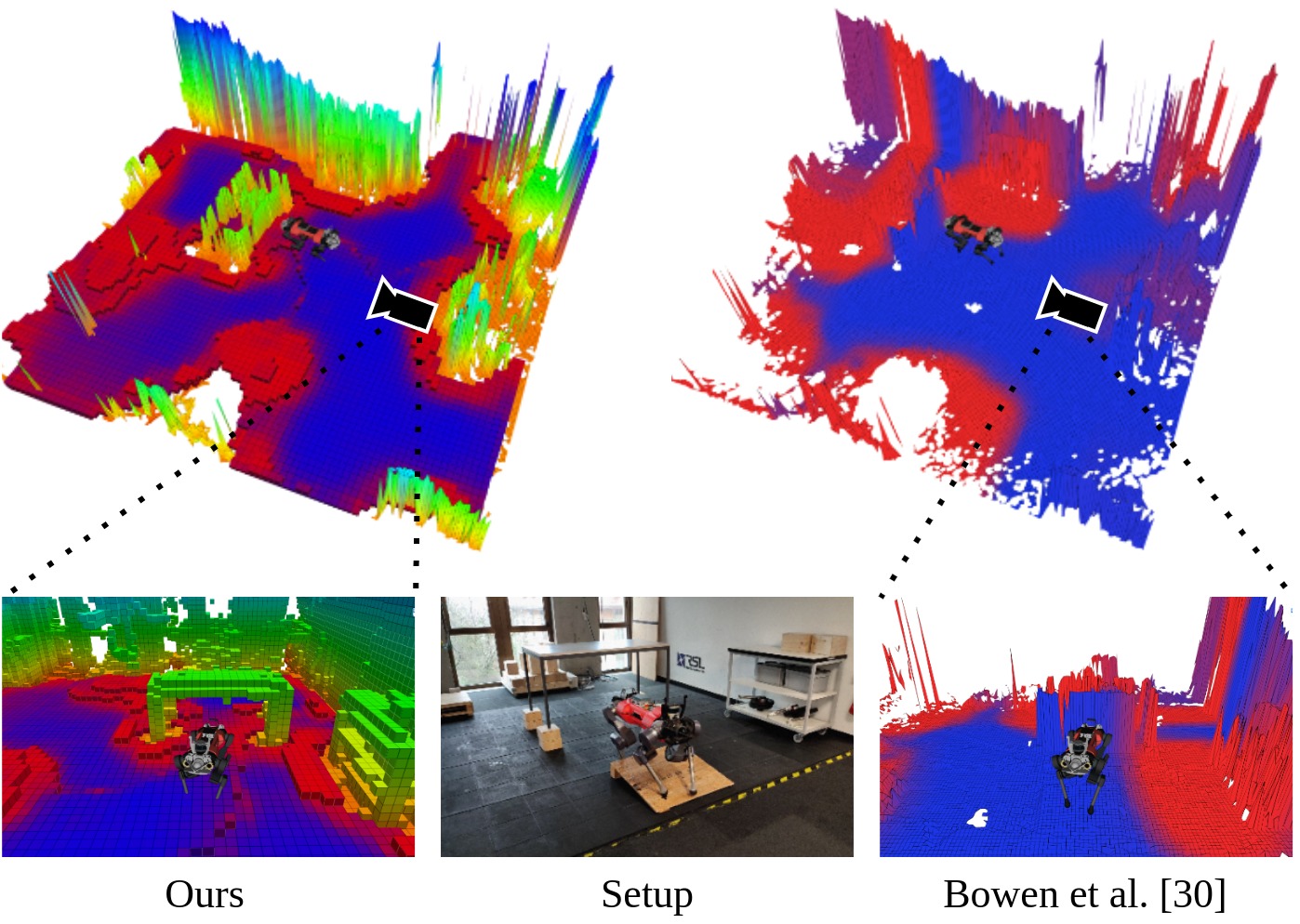}
\caption{Comparison of our network to an elevation mapping based traversability estimation approach. Our approach correctly labels the area underneath the desk as traversable. Elevation mapping based approach~\cite{Bowen21} is unable to produce a correct label for this area due to the corrupted elevation map.}
\label{fig:desk}
\end{figure}
\begin{figure}[ht]
\centering
\vspace{-0.5cm}
\includegraphics[trim=50 0 200 0,clip,width=\linewidth]{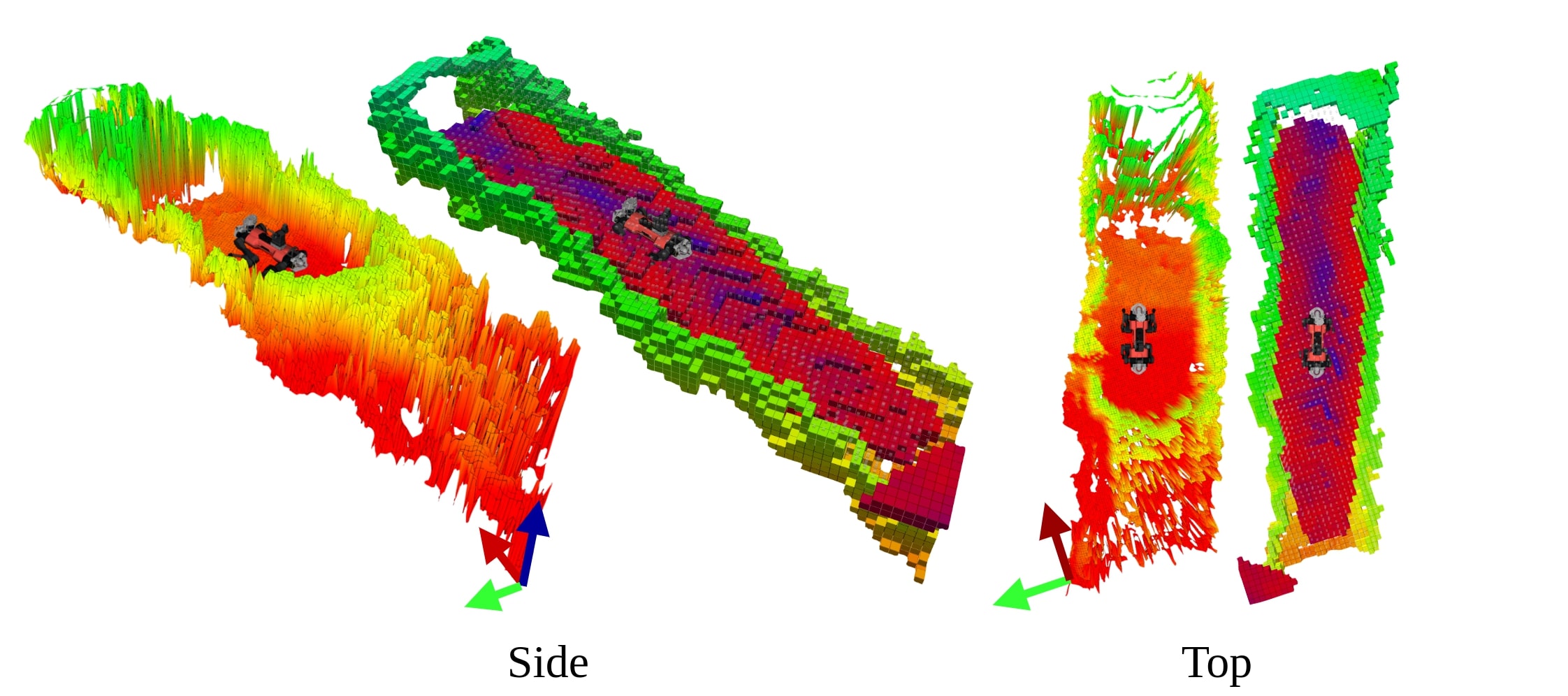}
\caption{Compromised elevation map in inclined low-ceiling mine shaft. Traversability prediction using the occupancy map. The ceiling is manually removed for visualization.}
\label{fig:mineshaft}
\end{figure}
\begin{figure*}[t]
\centering
\includegraphics[width=\linewidth]{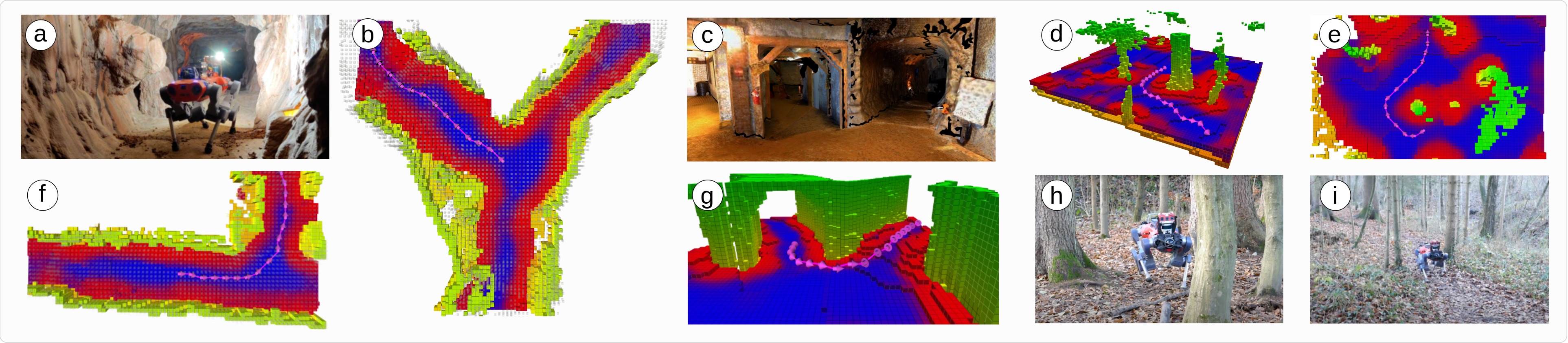}
\caption{Path planning in complex natural environment based on traversability estimate. Planned path is illustrated in purple. Forest (d,~e,~h,~i), Cave~(a,~ b,~c,~f,~g). In the cave images the ceiling voxels are manually removed for illustration purposes.}
\vspace{-0.5cm}
\label{fig:path_planning}
\end{figure*}
This experiment is designed to highlight the need for using a more expressive 3D volumetric environment representation than the typically used 2.5D elevation maps for traversability estimation, especially in the presence of overhanging obstacles.
Figure~\ref{fig:desk} shows the experimental setup with an elevated desk placed at the center of the room, allowing the ANYmal to pass underneath the desk without a special motion or gait.
The elevation and occupancy map are generated using the same LiDAR measurements and robot pose estimates provided by CompSLAM~\cite{compSLAM2020}. 
The elevation map is generated with a spatial resolution of \SI{5}{cm} on a Nvidia Jetson Xavier AGX~\cite{TakaElevationMapping}. 
The occupancy map is generated using VoxBlox~\cite{Helen17} with a spatial resolution of \SI{10}{cm} on the CPU. 
The occupancy map adequately captures the free space underneath the desk, while the elevation map of the same scene is unable to represent the area under the desk as free given that points on the tabletop are integrated.
We compare our work to the traversability estimation approach proposed in~\cite{Bowen21}, which uses an elevation map as an input. This network is trained on simulated data for ANYmal using a different data generation procedure. We use the estimated risk output of~\cite{Bowen21} to conduct a qualitative comparison solely. No quantitative results are presented given that both networks are trained on different data, and no ground truth information of the real traversability score can be obtained.
The network of \cite{Bowen21} is queried for each cell within the elevation map to assess the risk for 8 different headings when moving forward. In total 204.800 start-goal pairs per elevation map are evaluated and the average risk is color-coded per cell in Figure~\ref{fig:desk}. Red indicates untraversable terrain and blue risk-free traversable terrain. 
Given the corrupted input representation, the area underneath the desk is classified as non-traversable, which prohibits planning underneath the desk. 
As a comparison, the output of our \textit{total risk} network configuration is shown on the left in Figure~\ref{fig:desk} for the same scene. Here it is essential to distinguish between the color-coding used. In our work red implies traversable with a high risk (\cite{Bowen21} untraversable) and blue risk-free traversable (\cite{Bowen21} risk-free traversable). Our network predicts a traversability risk more adequately for the presented scene. Close to obstacles, the risk increases, and the occupied space is labeled as untraversable. 
Additionally, in Figure~\ref{fig:mineshaft} we illustrate a scenario of ANYmal traversing in an inclined mine shaft with a low ceiling. Due to low ceiling clearance the elevation map gets corrupted as depth measurements on the ceiling are integrated.
Our network, on the other hand, utilizing a 3D occupancy map can adequately predict the traversability cost. The prediction only includes minor errors (bottom left, predicting traversable region on top of the mine shaft), which can optionally be filtered out using a flood-fill algorithm. Here it is important to mention that scenarios like long steep inclined tunnels are not specifically covered within the training data. 
\subsubsection{Confined Spaces}
\label{subsubsec:exp_corridor}
\begin{figure}[ht]
\centering
\includegraphics[width=1\linewidth]{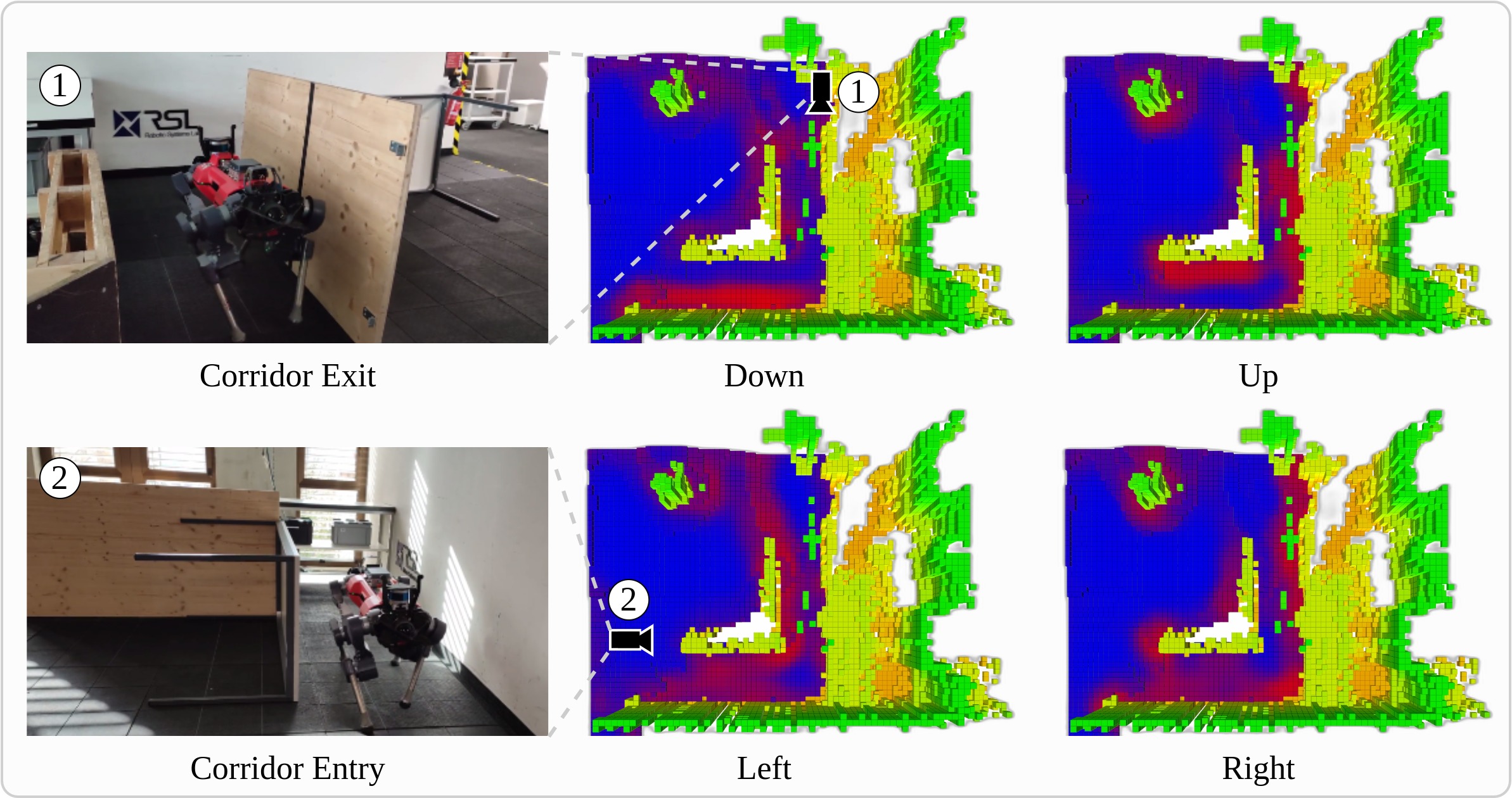}
\caption{Deployment of ANYmal within a confined corridor. Individual output channels for \textit{4-bin
motion direction risk} estimation.}
\label{fig:corridor}
\vspace{-0.5cm}
\end{figure}
In the following experiment, we deploy ANYmal within a confined space in an L-shaped corridor. The experimental setup is visualized from two view-points in Figure~\ref{fig:corridor}. 
The output of the network is illustrated for the \textit{4-bin
motion direction risk} configuration. The bins can be associated with moving up [\SI{315}{\degree},\SI{+45}{\degree}], right [\SI{45}{\degree},\SI{135}{\degree}], down [\SI{135}{\degree},\SI{225}{\degree}] and left [\SI{225}{\degree},\SI{315}{\degree}]. 
A higher risk is associated with actions that lead to a collision. 
Within the corridor entry, the cost of moving down and up strongly depends on the robot's position. 
This additional information can be helpful for safe path planning and allows to limit or rank motion commands at a spatial location within a terrain according to the excepted risk. 
During path planning this additional information can guide the search and may improve the planning performance in terms of risk or planning time. 

\subsubsection{Path Planning in Complex and Natural Environments}
\label{subsubsec:exp_natural}

To illustrate the real-world application of our proposed method, we deployed it on data gathered within a forest and the data from the DARPA subterranean challenges final run of team CERBERUS.
We use the \textit{total risk} network configuration to predict the traversability based on the generated occupancy map. 
Within the forest and underground cave, no non-rigid obstacles as e.g. high-grass or bushes, are present. Therefore the generated occupancy map representation correctly reflects the support surfaces. 

To showcase the application of our traversability estimation network we implement a simple local planner capable to find a safe path between the current robot position at the center and an externally provided target position.
Given the predicted traversability cost a directional graph is generated. Each traversable voxel is a node within the graph connected with its 26-neighborhood by an edge with a cost of $c_{s-g}$:
\begin{equation}
    c_{s-g} = \left\Vert \mathbf{p}_g - \mathbf{p}_s \right\Vert_2 + \lambda (1-\hat{\mathbf{T}}(\mathbf{p}_g)),
\end{equation}
where $\mathbf{p}_{s/g}$ denotes the position of the start and goal node. The hyper-parameter $\lambda$ trades off the length and total risk of the path. By increasing $\lambda$, lower total risks paths are preferred at the cost of a geometrical longer path between start and goal node.
The Dijkstra algorithm is used to find the shortest path between the current robot position and the provided target position. 
The resulting path for the forest and cave environment with $\lambda=0.1$ is illustrated in purple for all 5 example scenes in Figure~\ref{fig:path_planning}. 
Within the forest, the generated path traverses with a safety margin around all nearby trees. 
In the cave environment, the generated path traverses along the center of the corridor, maximizing the distance between both walls.  
Finally, we used the dataset collected within the forest to assess the average inference time of the network on a Nvidia Jetson Xavier AGX. 
The inference time depends on the number of occupied voxels given that computation is only performed in  a sparse manner using the Minkowski Engine and terrain geometry. 
On average, the network runs at \SI{16.5}{hz}$\pm$\SI{1.2}{hz} processing 12181$\pm$1611 input voxels. The inference time is measured over 100 selected key-frames in the forest dataset and the number of occupied voxels may change depending on the environment. Given the update rate of each LiDAR is \SI{10}{hz} the network is real-time capable.
\section{CONCLUSION}
In this work we showed that the proposed sparse 3D convolution neural network trained on traversability data collected in simulation on randomly generated terrains can accurately predict the traversability for unseen terrains using a fully geometric occupancy representations as input. 
Therefore the network trained exclusively on simulated data generalizes successfully to the real world. 
It successfully handles overhanging obstacles, outperforming elevation map based approaches in these scenarios, while being real-time capable. 
While we showed in multiple experiment successful sim-2-real transfers a more exhaustive study design of the sim-2-real transfer is needed. 
Not only a sim-2-real transfer of the actual traversability estimation network is needed but in addition the sim-2-real transfer of the deployed locomotion policy. 
For future work we will investigate integration of visual semantic information to assess friction coefficients or the quality of individual footholds. 
Additionally, no temporal consistency is enforced between individual predictions in the current implementation. This leads to changing predictions for individual voxels over time. 
Expanding the created robot-specific traversability estimation module to different robot categories will allow to develop a unified planning system operating across robot types without requiring robot-specific heuristics.
Finally, exploiting the generated traversability estimates for a variety of downstream tasks, including active switching between locomotion policies or gait patterns, local and global path planning, and exploration is promising. 

{\small
\bibliographystyle{IEEEtranS}
\bibliography{references}
}

\end{document}